\documentclass{llncs}
\begin{document}

\title{\textbf{Deep Recurrent Neural Networks for Product Attribute Extraction in eCommerce}}
\titlerunning{Bidirectional LSTM-CRF for eCommerce}  
\author{Bodhisattwa Prasad Majumder*, Aditya Subramanian*, Abhinandan Krishnan, Shreyansh Gandhi, Ajinkya More\\
(\textit{* denotes equal contributions})}

%
%
\institute{Walmart Labs}

\maketitle              

\begin{abstract}
Extracting accurate attribute qualities from product titles is a vital component in delivering eCommerce customers with a rewarding online shopping experience via an enriched faceted search.  We demonstrate the potential of Deep Recurrent Networks in this domain, primarily models such as Bidirectional LSTMs and Bidirectional LSTM-CRF with or without an \emph{attention} mechanism. These have improved overall $F_{1}$ scores, as compared to the previous benchmarks \cite{ajinkya2016} by at least 0.0391, showcasing an overall precision of 97.94\%, recall of 94.12\% and $F_{1}$ score of 0.9599. This has made us achieve a significant coverage of important facets or attributes of products which not only shows the efficacy of deep recurrent models over previous machine learning benchmarks but also greatly enhances the overall customer experience while shopping online.

\end{abstract}

\section{Introduction}
\subsubsection{\textbf{Problem Definition:}}

Facets allow e-commerce customers to narrow down a search space, for example by restricting size of clothing or screen size of televisions. An exhaustive and accurate coverage of facets can ensure a pleasurable and efficient navigation experience through the given product space \emph{iff} the attribute value metadata of the product properly appears in the facets for the given attribute. 

Take $S$ to be a search query entered by a user and $\mathcal{R}$
the set of products retrieved as result. Suppose product $\mathfrak{p}\in\mathcal{R}$, and $\mathfrak{p}$ has an attribute $\alpha$ which  happens to be a facet. Suppose further that the value of $\alpha$ applicable
to $\mathfrak{p}$ is $v$. When the user clicks on the attribute $\alpha$ with the associated facet value $v$, the filtered result set is $\mathcal{R}^{'}\subseteq\mathcal{R}$.
Then $\mathfrak{p}\in\mathcal{R}^{'}$ if and only if $\mathfrak{p}(\alpha)=v$. An example of the problem statement is defined here. 
	
    Let $x$ be a product title and let $(x_{1},x_{2},..,x_{n})$
be a particular tokenization $\mathbf{x}_{\mathbf{t}}$ of $x$. Given
an attribute $\alpha$, \emph{attribute extraction }is the process
of discovering the functions $E_{seq}$ (raw extraction) such that
\begin{itemize}
\item $E_{seq}(\mathbf{x_{t}})=E_{seq}((x_{1},x_{2},...,x_{n}))=(x_{i},x_{i+1},...,x_{k})$
for $1\leq i\leq k\leq n$ where $\mathbf{a}_{\mathbf{v}}=(x_{i},x_{i+1},...,x_{k})$
is a tokenization of a particular value of $\alpha$
\end{itemize}

\begin{example}
Consider the product title: 
\begin{center}
\begin{quote}
$\mathbf{x_{t}}=$ Hewlett Packard B4L03A\#B1H Officejet Pro
Eaio
\end{quote}
\end{center}
\medskip{}
Let $\alpha$ be the attribute `\textbf{\textit{Brand}}' we are interested in. Then we seek to
find a function $E$ (after whitespace tokenization) such that 
\[
E_{seq}((x_{1},x_{2},...,x_{6}))=(x_{1},x_{2})=(\text{Hewlett, Packard})  
\]
\end{example}

To achieve this solution, we implement a sequence to sequence labeling attribute extraction system which utilizes deep recurrent models at its core, including the Bidirectional LSTM Conditional Random Field with or without an attention mechanism. In traditional machine learning models, feature definition is an integral component to effective model performance \cite{ajinkya2016}. However, deep models are able to learn the present features, without being predefined, ultimately allowing the system to be flexible in approximating sequence learning functions through new attributes. Examining the differences between the brands extracted by the two models, Table \ref{tab:extract} shows a subset of the extraction discrepancies which are currently being targeted for corrections.
\begin{table}[h!]
\begin{centering}
\protect\caption{Attribute Extraction Performance for `\textit{Brand}' (Current vs. Previous Best \cite{ajinkya2016})}
\textsf{\scriptsize{}}%
\begin{tabular}{c|c|c}
\ChangeRT{1pt} 
\centering{\scriptsize{}\textbf{Product Title}} & {\scriptsize{}\textbf{Previous Best}} & {\scriptsize{}\textbf{Current Deep Model}}\tabularnewline
\ChangeRT{1pt} 
{\scriptsize{}\textbf{Woodland Imports} Decorative Bottle} & \scriptsize{}Woodland & \scriptsize{}\textbf{Woodland Imports}\tabularnewline

{\scriptsize{}\textbf{Home Essentials} White Essentials Sugar \& Creamer} & \scriptsize{}unbranded & \scriptsize{}\textbf{Home Essentials}\tabularnewline

{\scriptsize{}\textbf{Plum Island Silver} Sterling Silver Fairy Piece Ear Cuf} & \scriptsize{}Plum Island & \scriptsize{}\textbf{Plum Island Silver}\tabularnewline
\hline 
\end{tabular}
\par\end{centering}{\scriptsize \par}
\label{tab:extract}
\end{table}
%
%
\subsubsection{Labeling Scheme for Annotating Product Titles:} \label{subsec:annotation}
We describe our data annotation for training and validation purposes. In this example, we consider the attribute `\textit{Brand}'. The BIO
encoding scheme assigns one of the following three labels to each
of the tokens in the title.
\begin{enumerate}
\item \textbf{B-attribute}: Token is the beginning token of an attribute value \emph{iff} title contains the attribute value. 
\item \textbf{I-attribute}: Token is the intermediate token of an attribute value, if exists. 
\item \textbf{O}: Token is not representative of an attribute in the title. 
\end{enumerate}
To illustrate on a product title with the associated labels: 
`The Green Pet Shop Self Cooling Dog Pad'.

\medskip{}
$\begin{aligned}\underbrace{\text{The}}_{\text{B-attribute}}\ \underbrace{\text{Green}}_{\text{I-attribute}}\ \underbrace{\text{Pet}}_{\text{\text{I-attribute}}}\ \underbrace{\text{Shop}}_{\text{\text{I-attribute}}}\
\underbrace{\text{Self}}_{\text{\text{O}}} \ 
\underbrace{\text{Cooling}}_{\text{\text{O}}} \ 
\underbrace{\text{Dog}}_{\text{\text{O}}} \ 
\underbrace{\text{Pad}}_{\text{\text{O}}} \ 
\end{aligned}
$
\section{Model} \label{sec:methods}

\subsubsection{Long Short Term Memory (LSTM) Network:}
Recurrent Neural Networks (RNN) are built to understand contextual significance, but fall short of this task due to vanishing gradient problems wherein earlier parts of the network are less affected by backpropogation as compared to later parts of the network, resulting in convergence to suboptimal local minima \cite{bengio1994}. Long Short Term Memory (LSTM) networks, which were adaptations of vanilla RNNs were introduced to address this problem by implementing a forget gate layer and a memory cell \cite{lstm1997}. 
	
We use the following implementation, where $\sigma$ is the logistic function, $\odot$ is an element-wise product and $i$, $f$, $c$, $o$ and $h$ are the \textit{input gate}, \textit{forget gate}, \textit{cell}, \textit{output gate} and \textit{hidden} vectors respectively of same length. The weight matrix \textbf{W} has the corresponding subscripts for the different gates as the notation suggests. The LSTM cell generates a hidden vector $\textbf{h}_{t}$ which at every time step abstracts the previous context.
\begin{align*}
         & \mathbf{i}_{t} = \sigma(\mathbf{W}_{xi}\mathbf{x}_{t} + \mathbf{W}_{hi}\mathbf{h}_{t-1}  							+ \mathbf{W}_{ci}\mathbf{c}_{t-1} + \mathbf{b}_{i}) \\  
        & \mathbf{f}_{t} = \sigma(\mathbf{W}_{xi}\mathbf{x}_{t} + \mathbf{W}_{hi}\mathbf{h}_{t-1}  							+ \mathbf{W}_{ci}\mathbf{c}_{t-1} + \mathbf{b}_{i}) \\
         & \mathbf{c}_{t} = \mathbf{f}_{t} \odot \mathbf{c}_{t-1} + \mathbf{i}_{t} \odot tanh							(\mathbf{W}_{xc}\mathbf{x}_{t} + \mathbf{W}_{hc}\mathbf{h}_{t-1} + 							\mathbf{b}_{c}) \\
        & \mathbf{o}_{t} = \sigma(\mathbf{W}_{xo}\mathbf{x}_{t} + \mathbf{W}_{ho}\mathbf{h}_{t-1}  							+ \mathbf{W}_{co}\mathbf{c}_{t-1} + \mathbf{b}_{o}) \\
        & \mathbf{h}_{t} = \mathbf{o}_{t} \odot tanh(\mathbf{c}_{t})
\end{align*}

\subsubsection{\textbf{Bidirectional LSTM-CRF:}}

In previous work, notably Huang et. al. ~\cite{huang2015} proposed the usage of Bidirectional LSTM-CRF for named entity recognition by means of a sequence tagging task. 
Lample et. al. \cite{lample2016} also proposed a bidirectional LSTM with a sequential conditional random layer along with a new model of stack LSTMs with transition-based parsing. Similar models for sequence to sequence labeling models are frequently discussed in literature for various applications (\cite{seq2seq2013}, \cite{dyer2015}).

The Bidrectional LSTM has garnered a lot of attention as it takes into context both past and future tokens when understanding the current token at time $t$ ~\cite{grave2005}. Given the sequence of vectors $(\mathbf{x}_{1},\mathbf{x}_{2},...,\mathbf{x}_{n})$, the hidden vector serves as a concatenation of the hidden vectors from forward and backward states. If $\textbf{h}_{t}^{left}$ denotes the hidden vector obtained from forward flowing states and $\textbf{h}_{t}^{right}$ denotes the hidden vector obtained from backward flowing states, then the hidden representation of a token would be $\textbf{h}_{t}^{B-LSTM}$ = $[\textbf{h}_{t}^{left}; \textbf{h}_{t}^{right}]$.

\begin{figure*}[!h]
    \centering
    \includegraphics[width=80mm]{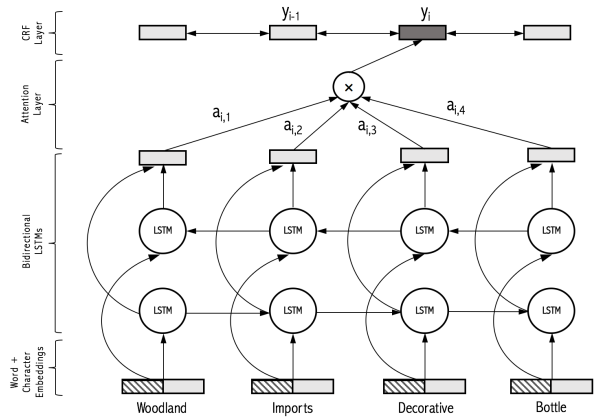}
    \caption{Representation of Bidirectional LSTM-CRF Attention Model}
    \label{fig:models}
\end{figure*}

The Conditional Random Field (CRF) is a probabilistic structured prediction model which predicts future labels, while taking into account previously predicted labels as well \cite{lafferty2001}. Lample et. al. \cite{lample2016} and Huang et. al. \cite{huang2015} both showed the advantages of CRF for label tagging using the features generated via the Bidirectional LSTM network as input parameters. An LSTM-CRF effectively uses LSTM layers to capture contextual information from the input sequence and a CRF at the output layer for efficient label tagging. This removes the need to hand engineer features for the CRF to learn. The CRF layer is learned by optimizing the parameters in the state transition matrix for the tags. Let's assume $\mathbf{M}$ is the matrix of scores given by the Bidirectional LSTM Network where $\mathbf{M}_{i}^{j}$ is the score of the $j$-th tag for the $i$-th token of the sequence. For a predicted sequence $(y_{1}, y_{2},..., y_{n})$, where $\mathbf{A}$ is the transition matrix for the tags, the combined score would be 
\[
s(\mathbf{x}, \mathbf{y}) = \sum_{i=0}^{n} \mathbf{A}_{y_{i}}^{y_{i+1}} + \sum_{i=1}^{n} \mathbf{M}_{i}^{y_{i}}
\]

Finally, for all the methods described above, a softmax over all possible tags would provide the probabilities for the output tag sequence. The log-probability of the correct tag sequences is to be maximized during training \cite{lample2016}. Figure \ref{fig:models} captures the diagrammatic representations of the Bidirectional LSTM-CRF architecture. 

\subsubsection{\textbf{Attention Mechanism:}}
The attention mechanism \cite{olah2016} allows for an LSTM network to isolate tokens of contextual and locational interest from both past and future indices. Contextual information is used to understand tokens useful to the current index, and utilize those. Locational information complements this by allowing tokens to move around in memory, enabling the attention mechanism to persist through the entire network. As visualized in Figure 1, value for $a_{i,j}$ corresponds to respective attention weight prescribed by the $j$-th initial word token for output token $y_i$. 

\subsubsection{\textbf{Word and Character Embeddings:}}
A word embedding is a lower dimensional dense representation of a word which encodes not only the intrinsic meaning of the word but also the semantic meaning given its usage in various contexts. On the other hand, character embeddings \cite{charembed} encapsulate patterns not explicitly noticeable through words. This for instance helps  capture `\textit{Brand}' attributes, in the case that the given brand does match with a respective word embedding. We consider both the word and character embeddings as random vectors for the initial run, but later allow the network to learn the embeddings based on the data and the task by means of an embedding layer.

\section{Results} \label{sec:results}
\subsubsection{Dataset and Training Setup:} \label{subsub:exp}

We obtained product titles from online catalogs containing a variety of products. Our experiments pertaining to this paper concentrate around the attribute `\textit{Brand}'. 
For `\textit{Brand}', we collect $61,374$ product titles for the experiment. Training, validation, and test data are generated with a 60/20/20 split ratio respectively. Titles are further tokenized by whitespace and labeled according to the annotation scheme described in Section 1. For accurate labels to train and validate our model, we acquire `\textit{Brand}' attributes for the set of product titles through crowdsourcing tasks. 

We employ stochastic gradient descent as a learning method to allow the gradient to back-propagate through time (BPTT). For all the deep models we consider word embeddings of size 100 and character embeddings of size 25. We add dropout layers with dropout rate 0.2 \cite{dropout}, and all models were run for 200 epochs with 5-fold cross validation. Higher dropout rate negatively affected our results.


\subsubsection{Model Performance:}
Bidirectional LSTM with CRF layers outperforms all the other methods attempted as can be seen from Table 2.
The $F_{1}$ measure for the Bidirectional LSTM-CRF is \textbf{0.9599} which is highest among all deep models. Compared to the previous best two models \cite{ajinkya2016}, Structured Perceptron and Linear Chain Conditional Random Field, the $F_{1}$ scores rose by \textbf{0.0392} and \textbf{0.0391} respectively. Even though attention did not positively impact the $F_{1}$ score for the Bidirectional LSTM-CRF, attention improved precision when applied over a Bidirectional LSTM without the CRF layer at the output.

\begin{table}[h!]
\begin{centering}
\protect\caption{Model metrics for `\textit{Brand}' Extraction with 5-fold Cross Validation}
\textsf{\scriptsize{}}%
\begin{tabular}{l|c|c|c|C{1.8cm}}
\ChangeRT{1pt} 
& \textbf{\scriptsize{}Precision(\%)} & \textbf{\scriptsize{}Recall(\%)} & \scriptsize{}$\mathbf{F_{1}}$\textbf{-Score} & \textbf{\scriptsize{}Label Accuracy(\%)}\tabularnewline
\ChangeRT{1pt}
\textbf{\scriptsize{}Bidirectional-LSTM-CRF} & \scriptsize{}\textbf{97.94} & \scriptsize{}\textbf{94.12} & \scriptsize{}\textbf{0.9599} & \scriptsize{}\textbf{99.44}\tabularnewline

\textbf{\scriptsize{}Bidirectional-LSTM-CRF Attention} & \scriptsize{}97.38 & \scriptsize{}93.98 & \scriptsize{}0.9565 & \scriptsize{}99.44\tabularnewline

\textbf{\scriptsize{}Bidirectional-LSTM Attention} & \scriptsize{}95.12 & \scriptsize{}92.80 & \scriptsize{}0.9395 & \scriptsize{}98.92\tabularnewline

\textbf{\scriptsize{}Bidirectional-LSTM} & \scriptsize{}92.16 & \scriptsize{}92.72 & \scriptsize{}0.9244 & \scriptsize{}98.92\tabularnewline
\hline 

\textbf{\scriptsize{}Structured Perceptron \cite{ajinkya2016}} & \scriptsize{}91.98 & \scriptsize{}92.18 & \scriptsize{}0.9208 & \scriptsize{}98.44\tabularnewline

\textbf{\scriptsize{}Linear Chain Conditional Random Field \cite{ajinkya2016}} & \scriptsize{}91.94 & \scriptsize{}92.21 & \scriptsize{}0.9207 & \scriptsize{}98.44\tabularnewline
\hline

\end{tabular}
\par\end{centering}{\scriptsize \par}
\label{tab:results}
\end{table}
\subsubsection{Discussion:}
Using the attention mechanism does not guarantee higher precision for the Bidirectional LSTM-CRF. We speculate this being the case as the primary usage of a Bidirectional LSTM in this application is to deliver features to the CRF. The CRF is the final layer of the model which delivers an attribute label. The attention mechanism seems to interfere with the ability of the CRF layer to extract the attribute in question. However, in the absence of a CRF, with a Bidirectional-LSTM model, attention helps the model performance, indicating that the attention mechanism and the CRF layer are both likely performing a similar function.

\section{Conclusion} \label{sec:conclusion}

The attribute extraction system has significant impact on the product discoverability in a faceted search system which can be estimated from the online impressions on the products after a new set of attribute values have been added. In addition to `\textit{Brand}' whose addition shows considerable improvement on clicks, add to cart rate and orders, other attributes also where attribute extraction was effective show a positive impact on impressions. `\textit{Internal Memory}', `\textit{Product Line}', and `\textit{Manufacturer Part Number}' are examples of such attributes with 3.02\%, 1.02\% and 4.38\% increases in precision scores respectively. Furthermore, to measure the impact, we conducted an experiment with a set of 272,697 products where approximately 250,000 additional impressions per day were observed over a period of 27 days after brand attribute values were added.  This improvement in impressions is a result observed on the previous system. While our current method achieves 0.0391 increase in the $F_{1}$ score over the previous best methods, replacing previous models by the deep model potentiates a significant boost in impressions and other key metrics indicating notable business impacts.

\end{document}